\documentclass{bmvc2k}

\title{Gaussian map predictions for 3D surface feature localisation and counting}

\addauthor{Justin Le Louëdec}{jlelouedec@lincoln.ac.uk}{1}
\addauthor{Grzegorz Cielniak}{GCielniak@lincoln.ac.uk}{1}

\addinstitution{
 Lincoln Centre for Autonomous Systems,\\
 University of Lincoln, \\
 Lincoln  LN6 7TS, \\
 United Kingdom
}

\runninghead{Le Louëdec and Cielniak}{Gaussian map predictions for 3D surface...}

\usepackage{graphicx}
\usepackage{mathtools}
\usepackage{tabularx}
\usepackage{url}
\usepackage{multirow}
\usepackage{siunitx}
\begin{document}

\maketitle

\begin{abstract}
In this paper, we propose to employ a Gaussian map representation to estimate precise location and count of 3D surface features, addressing the limitations of state-of-the-art methods based on density estimation which struggle in presence of local disturbances. Gaussian maps indicate probable object location and can be generated directly from keypoint annotations avoiding laborious and costly per-pixel annotations. We apply this method to the 3D spheroidal class of objects which can be projected into 2D shape representation enabling efficient processing by a neural network GNet, an improved UNet architecture, which generates the likely locations of surface features and their precise count. We demonstrate a practical use of this technique for counting strawberry achenes which is used as a fruit quality measure in phenotyping applications. The results of training the proposed system on several hundreds of 3D scans of strawberries from a publicly available dataset demonstrate the accuracy and precision of the system which outperforms the state-of-the-art density-based methods for this application.
\end{abstract}

%-------------------------------------------------------------------------
\section{Introduction}
\label{sec:intro}

Recent advances in computer vision have led to several practical applications addressing challenging problems in fields such as agriculture. Modern imaging systems enable rapid, non-destructive monitoring of different crops and their traits, often referred to as high-throughput phenotyping, which are useful for breeders and biologists~\cite{yang2020crop}. The current state of the art in object detection, which might be used for such tasks, targets entire objects or their large parts (e.g.~\cite{he2017mask,pascal-voc-2012,10.5555/3295222.3295263,shi2021pv}) and are therefore not suitable for extracting small and precise regions which might be vital for characterising properties of such objects. The shape information is of particular significance in phenotyping applications and therefore 3D vision can bring several benefits when compared to 2D images. 

Standard supervised segmentation and detection algorithms require very tedious annotation process, which is very complex and difficult to acquire for small surface detail of 3D objects.
An alternative and simpler approach is to employ keypoint annotations which denote the central location of each object and use this information to create density maps. Such annotations were successfully employed for counting by regression through prediction of density maps has been proposed \cite{2018arXiv180210062L,2018arXiv181110452L}. Such methods are very efficient in predicting object counts across images and areas of interest where a high density of objects or details can be observed. A drawback of such methods, however, is their lack of precision for localising the counted details.

In our work, we consider a particular application of strawberry achene counting which is an important characteristics of fruit sought by both strawberry producers and breeders, making it an important phenotyping task. The fertilised achenes, or the visible ``seeds'' on the surface of a berry, are responsible for the development of the fruit and its overall quality. The achenes are arranged into fairly regular spiral rows and their number is affected by the cultivar, berry's location on the plant and environmental factors~\cite{khanizadeh1994automated}. The counting process is typically manual, laborious and therefore there is a great interest in pursuing automated methods for undertaking this task. Some earlier attempts looked at the use of 2D colour imagery for that purpose~\cite{he2017novel,li2020defining}, but due to a large variation in achene's colour and appearance such methods are suffering from limited accuracy. Therefore we propose to exploit the use of 3D information for this task.

In contrast to state of the art, our work proposes to employ a Gaussian map representation to estimate precise location and count of 3D surface features together with a custom-made network trained on simple keypoint information indicating central location of achenes in image coordinate, addressing limitations of the density-based methods. The contributions of this paper are as follows: 1) a method for utilising the 3D spheroidal nature of certain objects for creating 2D surface projections for efficient processing by Convolutional Neural Networks (CNNs), 2) a custom-made neural network GNet trained on Gaussian maps for an accurate and efficient prediction of the surface locations and their precise count, 3) experimental comparison of the proposed method to state-of-the-art density-based approaches in a practical application of achene counting in 3D scans of strawberries.

\section{Related work}\label{sec:related_work}

\textbf{Surface analysis }of the objects rather than their spatial information, is a core idea behind our work. In~\cite{10.1145/3072959.3073616}, the authors propose a new method for applying deep learning over sphere-like objects using a parameterisation known as a planar flat torus. In~\cite{2018arXiv180804952Y}, the authors work on learning information on mesh surfaces which, due to their nature, better capture detail and geometry compared to point clouds or voxels, but with a non-Euclidean structure and non-trivial topology, the complexity associated with their processing rises significantly. They propose operations similar to convolutions over the surface to learn features for segmentation or classification. 
Another approach presented in~\cite{2018arXiv181210705H} is using a surface to image projection via torus, which achieves state-of-the-art performance on 3D object classification and segmentation. Mesh-based methods cannot be easily applied to scenarios where fine shape detail is required, due to the exponential complexity associated with rapidly increasing face-count in high precision meshes. In~\cite{cohen2018spherical}, the authors introduce a set of building blocks for CNN to use spectral information from spherical projection and Fourier transform applied to 3D model recognition. In~\cite{2017arXiv171204426C}, a similar approach is proposed but instead using the projections directly, with stripes along the azimuthal coordinates and increasing contour information with multi-view branches to the proposed framework.
Further analysis of stereographic projections and their suitability for the 3D object classification task is considered in~\cite{2018arXiv181101571Y}. Another approach to spherical data is presented in~\cite{deepsphere_iclr,deepsphere_cosmo,deepsphere_rlgm} which uses pooling and sampling over a graph created on the surface of a sphere, to learn classification and possibly other tasks. The promising results were observed in cosmology classification. The use of spherical and stereographic projections, however, was mainly restricted to classification tasks. Their uses for detection and segmentation are limited due to lack of suitable annotated datasets.

\textbf{Counting by regression} rather than by detecting each object is a strong alternative when exact localisation is not needed. In~\cite{NIPS2010_fe73f687}, the authors proposed a supervised framework for predicting the number of objects by using a loss functions based on the MESA distance and global density prediction. This work highlights the fact that density is not always being equal to the number of objects, especially at the border of images where the distribution pattern can be cut off. But this behaviour was seen as desirable, as objects on the sides of images where not fully counted. This density prediction is later used in~\cite{2018arXiv180210062L}, where different deep learning architectures where employed to predict density maps and use the same integration method to predict object counts. 
In~\cite{gao2021crowd}, the authors propose to combine Gaussian maps and density maps across different losses, to refine predictions of the different network parts. While improvements in counting performance compared to state-of-the-art methods are reported, the method still relies on density maps to predict the count, and requires complex training with multiple networks and loss functions involved. Improvements to previous detection frameworks on scenes with a high number of similar and closely located objects was reported in~\cite{kant2020learning} by using Gaussian maps as auxiliary predictions. While improving the detection and count results such a method still relies on fully annotated detections.

\textbf{In agriculture}, one of the important phenotyping measures is based on counting particular elements of plants. A comparison between two detection algorithms to produce a predicted number of leaves for a given plant with a rudimentary robotic platform is presented in~\cite{s20236896}. A similar idea is pursued in~\cite{dobrescu2019understanding} where the number of leaves is  regressed through a neural network's latent space  without producing any density map or similar representation for localisation. In~\cite{ubbens2020autocount}, unsupervised learning is used to provide pseudo-segmentation maps and several optimisation steps such as watershed segmentation are used to provide the object count, although the segmentation and count suffer from imprecision due to the unsupervised nature of the method. Counting object in agriculture for phenotyping purposes is such an important task that~\cite{9364677} offered a comparison of detection- and regression-based for various fruit types. In~\cite{lottes2018joint} localisation prediction combined with segmentation is used to refine the prediction for plant detection. The approach creates blobs for each object of different classes and employs one decoder branch for prediction and one for pixel segmentation. The method, however, still relies on segmentation annotation and the two decoder branches double the number of parameters for the network. A similar work is presented in~\cite{zabawa2019detection} for grape counting in images, focusing on using segmentation masks with 3 different classes to produce the count using connected components algorithm for clustering. But such a method requires precise annotation for fruits, background and fruits edges, which is difficult to obtain for large datasets.

\section{Method}

\subsection{Spherical projection to a planar representation}\label{sec:proj}

In this work, we consider a special class of spheroid-like objects, which are characterised by their shape similar to a deformed spheroid centred around their centre of mass. This characteristics allows for converting meshes of such objects from Cartesian to spherical coordinates, which can be then unrolled and used as indices of the 2D planar shape representation. This representation allows the use of 2D convolutions over the surface of the object and acts as straightforward dimensionality reduction.

Let's define $\mathcal{P}$ as the point cloud formed of $N$ points sampled over the surface of the object and assume that $\mathcal{P}$ is centred around the origin and aligned along the Z axis. Each point $P$ is represented in Cartesian coordinates as $(x,y,z)$ and in spherical coordinates as $(\rho,\theta,\varphi)$ where $\rho$ the radial distance from the origin/centre of mass, $\theta$ the latitude, and $\varphi$ is the polar angle. Coordinates in spherical coordinates are calculated as follows: $\rho=\sqrt{P_{x}^{2} +P_{y}^{2} +P_{z}^{2}}$, $\theta=\arctan({\sqrt{P_{x}^{2} +P_{y}^{2}}/P_{z}})$ and $\varphi=\arctan{(P_{y}/P_{x})}$. The shape in spherical coordinates is then sampled in $\Delta$ degree angle increments along the $\varphi$ in range of $[-180^{\circ},180^{\circ}]$ and $\theta$ in range of $[0^{\circ},180^{\circ}]$ resulting in a 2D representation $\mathcal{X}$ of $360/\Delta$ width and $180/\Delta$ height. For a complete coverage and to reduce the sparsity of the projected points, we use a cubic interpolation between the pixels of $\mathcal{X}$ to fill in any missing values. We use surface normals' components as the value of each pixels. These normals are obtained directly from the original mesh (with underlying $\theta$ and $\phi$ kept through the projection).

\begin{figure*}[!t]
\begin{center}
\begin{tabular}{c|c}
     \includegraphics[width=0.16\linewidth]{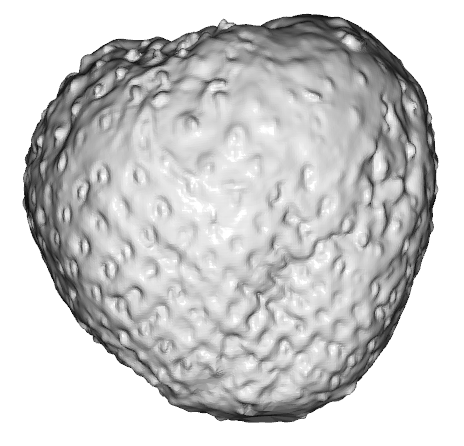}&
     \includegraphics[width=0.30\linewidth]{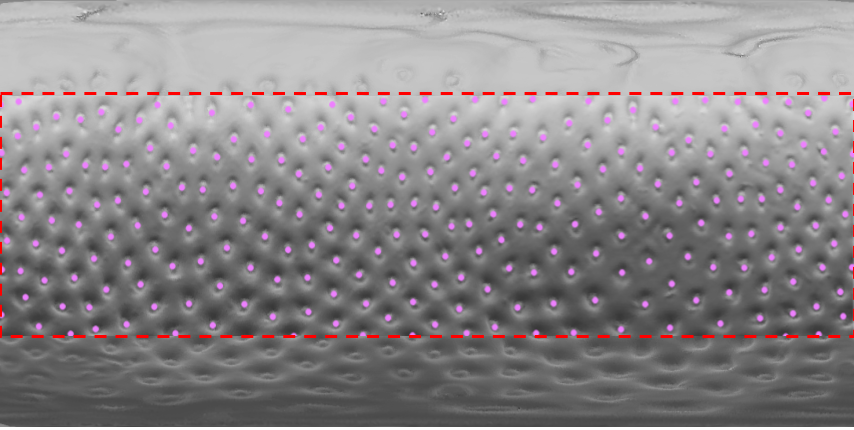} \\

\end{tabular}
\end{center}
   \caption{A 3D scan of strawberry from~\cite{straw_3d_data} (left) and its 2D projection with an indicated ROI and the annotated surface (right). The visualisation features surface normals along the Z axis.}
\label{fig:example1}
\end{figure*}

To illustrate the projection steps, Fig.~\ref{fig:example1} demonstrates an example from our application featuring a 3D strawberry scan from~\cite{straw_3d_data} and its projection onto the 2D space. Since, in this particular case, the information around both poles is of limited use (i.e. it represents removed calyx and clamped tip of the strawberry), we further restrict the region of interest in the projected image and limit the height to $h_{min}$ and $h_{max}$ values.

\subsection{Counting and localisation}

\begin{figure}[!t]
    \begin{center}
    \begin{tabular}{c}
      \includegraphics[origin=c,width=0.9\linewidth]{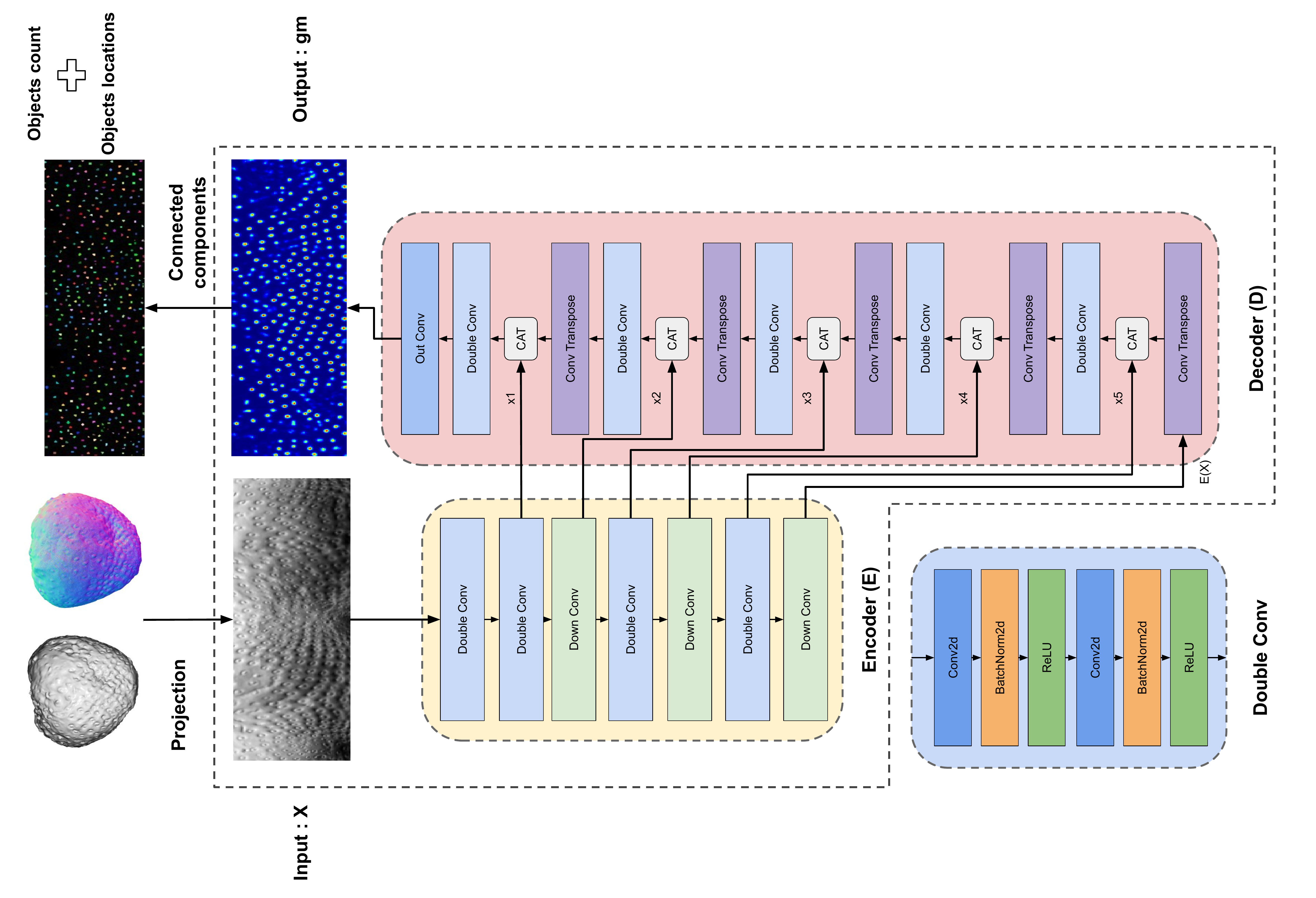}\\ %\includegraphics[origin=c,width=0.2\linewidth]{images/fullmethod.pdf} \\
    \end{tabular}
        
  \end{center}
    \caption{The complete pipeline for object localisation and count with the proposed GNet architecture.}
    \label{fig:network}
\end{figure}

We use supervised learning based on a 2D CNN to localise and count the surface features. Rather than using a traditional segmentation/detection model for each feature, we apply a regression approach with locations indicated by a Gaussian kernel centred around each surface feature which we want to count. The overall diagram indicating critical components of the method is presented in Fig.~\ref{fig:network}.

\textbf{Gaussian maps:} The ground truth Gaussian map is created from keypoint annotations indicating location of each surface feature (i.e. strawberry achene in our case) in the planar projection image. For each annotation, a symmetric 2D Gaussian is applied representing the likelihood of the pixels to be part of the object, with the highest pixel value at the keypoint location. Each Gaussian map $M_{G}$ can be described as $M_{G} = \sum_{i=1}^{N}\textbf{I}_0 +\textbf{G}_{\sigma}(i)$ for N annotated objects, where $\textbf{I}_0$ is an empty image corresponding to the size of the 2D projection $\mathcal{X}$, and $\textbf{G}_{\sigma}(i)$ is a Gaussian kernel centred around each annotation. The values of the final map are normalised so that the highest pixel value is 1.0 representing the exact and more likely location of the achene with radially decreasing confidence values. Contrary to the density maps, this representation indicates location of individual objects rather than their count. 

\sloppy In our work, we consider two selection criteria for $\sigma$: a fixed value for all objects, selected low enough to avoid overlaps in dense regions, and an adaptive value $\sigma_a=min(d_{min}p_t,\beta)$ based on the distance to the closest neighbour $d_{min}$ and binary likelihood threshold $p_t$ (described below) with a fixed upper bound $\beta$ corresponding to the size of an average object.

\textbf{Gaussian network (GNet):} We propose to use a supervised method to train a object localisation predictor with the generated ground truth Gaussian maps. The input image $\mathcal{X}$ consists of 3 channels corresponding to the normalised surface normal attributes. Our model is based on an encoder-decoder architecture derived from the UNet framework. The encoder $E(\mathcal{X})$ generates a latent feature space with $x_1,x_2,x_3,x_4,x_5$ different level/scale of the feature map. The output from the network $gm=D([x_1,x_2,x_3,x_4,x_5,E(\mathcal{X})])$ represents the latent space progressively upsampled and decoded through the decoder $D$, with the different scales of feature maps concatenated at different layer outputs. The graphical overview of the architecture is presented in Fig.~\ref{fig:network} which includes double convolution blocks (Double Conv) and combined maxpool-convolution blocks (Down Conv)\footnote {The source code is available at: \url{https://github.com/lelouedec/PhD_3DPerception}}. The key difference with the UNet framework resides in the upsampling blocks, which we have been replaced by transposed convolutions to allow the decoder to learn the progressive upsampling of feature maps and increase its precision. We also added dilation to the transposed convolutions, to spread apart the learned kernels across larger areas and improve the field of view and understanding of local information when upsampling. We offer an ablation study of these improvements in Sec.~\ref{sec:results}.

As loss function, we use the Binary Cross Entropy expressed as $L =\frac{1}{N}\sum_{i=1}^{N}-(y_{i}\log(gm_{i}) + ( 1 - y_{i})  \log(1-gm_{i}))$ for all $N$ pixels, where $gm_{i}$ is the predicted Gaussian map value and $y_{i}$ is the ground truth value at location $i$. The loss function in this form is traditionally used for binary classification problems, but it can also be used for measuring the distance between two probabilistic distributions.

\textbf{Object count from the predicted maps:} From the calculated Gaussian maps, the final object count is calculated as follows. First the output values are binarised using a fixed threshold $p_t$ which creates distinct groups of foreground pixels.
Then, the resulting binary map undergoes a connected component labelling procedure resulting in a set of individual clusters corresponding to the object count. 
The localisation information of the achenes, if required, can be obtained directly from the cluster centres, which is not possible to obtain directly when using density based methods.

\subsection{Baselines}

As a baseline for comparisons to our method, we consider a basic non-maxima suppression algorithm based on point distances from the origin ($\rho$) and two state-of-the-art approaches based on density estimation.

\textbf{Non-maxima suppresion method:} We create region proposals as circles of diameter $\beta$, located on local-maxima of the projection of $\rho$ on the strawberry surface. We then use non-maximum suppression over the proposed locations of achenes and combine into one the predictions with an overlap above 50\%. The final count of achenes is the number of  all predictions left after the suppression procedure.

\textbf{Density method:} Generating the density map can be realised by convolving each keypoint annotation with a Gaussian kernel resulting in a groundtruth density map. Each individual kernel is normalised so that all kernels can be integrated into the toal number of objects in the image. This idea is popular for example in crowd counting applications (eg.~\cite{2018arXiv181110452L}) which typically use loss functions based on distance in metric space (Mean Square Error loss). 

The density map is expressed as $M_{D}(x) = \sum_{i=1}^{N}\delta(x - x_i)G_{\sigma}(x)$, where $\delta(x - x_i)$ is a map with 1 at each location and 0 elsewhere and $G_{\sigma}(x)$ is a Gaussian kernel with $\sigma=\frac{\sum_{j=1}^{3}k_{j}}{f}$, where $k_{j}$ is the distance from the annotated pixel $i$ to the $j$th closest neighbour, and $f$ is a scaling factor. The final count of objects $C$ is obtained by integrating all density pixels from the entire image.

The two state-of-the-art baseline network architectures for density-based predictions are CSRNet~\cite{2018arXiv180210062L} and CAN~\cite{2018arXiv181110452L}. We chose these two methods, since their complexity is comparable to our model. CSRNet is a single column model with a VGG16~\cite{simonyan2014very} feature extractor as an encoder,  a mirrored decoder without upsampling and an outer convolution generating the predicted density map. We use version of the network with a dilation rate of 2 from the original paper. The CAN approach is multi-scale, using different feature maps from the encoder which are upsampled and combined into one latent space feature map before being fed to a decoder similar to one in CSRNet. We chose both methods for being the best performing methods for density based count regression on multiple standard datasets.

For both methods, the output size of the network is 8 times smaller than the input. Therefore, the target of the loss function needs to be downsampled using a bicubic function for training. This impacts the precision of the methods with smaller resolutions of inputs, and tends to force averaging over some areas instead of very precise density information.

\section{Evaluation}

\subsection{Datasets}

For our application, we use a publicly available dataset of high-resolution 3D scans of strawberries presented in~\cite{straw_3d_data}. The dataset comprises of images, 3D scans and phenotypic information for 1611 strawberries, divided into 20 groups representing 15 varieties for 3 different locations. The calyx was manually removed while trying to minimise the amount of discarded flesh. The 3D scans were performed with use of a high resolution Solutionix Rexcan DS2 scanner typically employed for precise dental scanning applications, and spray coating each berry with a white titanium solution to minimise light reflection. For each berry, 10 different views were captured to create the resulting point cloud which then underwent the Poisson disc reconstruction process using three depths of 4, 6 and 8. In our work, we chose the depth of 8 for the highest level of detail.

For the purpose of strawberry achene counting, we manually indicated keypoint locations for each achene in the images resulting from planar projections of the 3D scans (see Fig.~\ref{fig:example1} for example annotation). We annotated 781 strawberries in total corresponding to 11 groups of the original dataset including 7 different species and 4 different picking locations. For each strawberry, we generated both density and Gaussian maps as groundtruth for the different methods evaluated. We use an 80\% training and 20\% testing set split corresponding to 625 strawberry scans for the training set and 156 for the testing set. The training set has an average of 274 achenes (std: 66, min: 99, max: 503), and the testing set has an average of 271 achenes (std: 62, min: 98, max: 499).

\subsection{Experimental parameters}\label{sec:expe}

In our experiments, we compare four different methods which we denote as follows: GNet is the proposed method with a fixed Gaussian kernel, GNet$_a$ with a variable kernel and two baseline methods are denoted as CSRNet and CAN. The specific parameters for all four methods are detailed below.

\textbf{Input resolution:} We use two different resolutions for input $\mathcal{X}$ determined by a degree angle increment parameter which we set to $\Delta=0.5$ and $\Delta=1.0$ degrees, resulting in two image resolutions of $720\times360$ and $360\times180$. Based on various examples from the dataset, we choose the ROI over our projections to be at $h_{min}=23.5\%$ and $h_{max}=76.5\%$ allowing for ignoring areas previously covered by the calyx at the top and clamped flesh at the bottom. For the binarisation of the predicted Gaussian map $gm$, we use a threshold value $p_t=0.33$ giving $>66\%$ likelihood for the object to be found at the given location.

\textbf{Training parameters:}
To take into account potential rotation of the berries along the z axis, as well as avoiding too fast convergence toward local minimum and overfitting, we use data augmentation during training. To imitate the rotation around the Z axis, the projected training image is translated horizontally by a random value from a range [0, W]. This mainly helps the network generalising for the objects location. We use the Adam optimiser, with a learning rate of \num{1e-5} for density based methods and \num{1e-6} for GNet. All models are trained until the loss starts to plateau (loss oscillating around an average value). We train on a PC equipped with an NVIDIA 1080 Ti  GPU and 12 GB of VRAM. For $\Delta=1.0$, a single inference of GNet with connected component clustering performs at 25 FPS which drops down to 5 FPS for $\Delta=0.5$.

\textbf{Gaussian kernel:} For the GNet with a fixed kernel size, we set $\sigma=1.25$ for $\Delta=1.0$ and $\sigma=2.5$ for $\Delta=0.5$. For the adaptive variant GNet$_a$, we set the bounding value $\beta=2.5$ for $\Delta=1.0$ and $\beta=5$ for $\Delta=0.5$. These values are based on the average radius of the achenes which for the resolution corresponding to $\Delta=0.5$ equates to about 5 pixels. The scaling factor $f$ for the density-based methods CSRNet and CAN is set to 10 following the original implementation of the methods.

\textbf{Evaluation metrics:}\label{sec:Metodores}
Following previous evaluation methods for counting, we use as metrics the Mean Absolute Error (MAE) $\frac{\sum_{i=1}^{n} |y_{i}-x_{i}|}{n}$ and Root Mean Squared Error (RMSE) $\sqrt{\frac{\sum_{i=1}^{n} (y_{i}-x_{i})^{2}}{n}}$, where $y_{i}$ is the ground truth count for sample $i$, $x_{i}$ is the predicted count and $n$ the number of samples in the test set. We additionally provide the average percentages of False Positive (FP) and False Negative (FN) cases, to reflect better on the performance of each method. 

%-------------------------------------------------------------------------
\section{Results} \label{sec:results}

\begin{figure}[!ht]
\begin{center}
    \begin{tabular}{c|c}
        \includegraphics[width=0.49\linewidth]{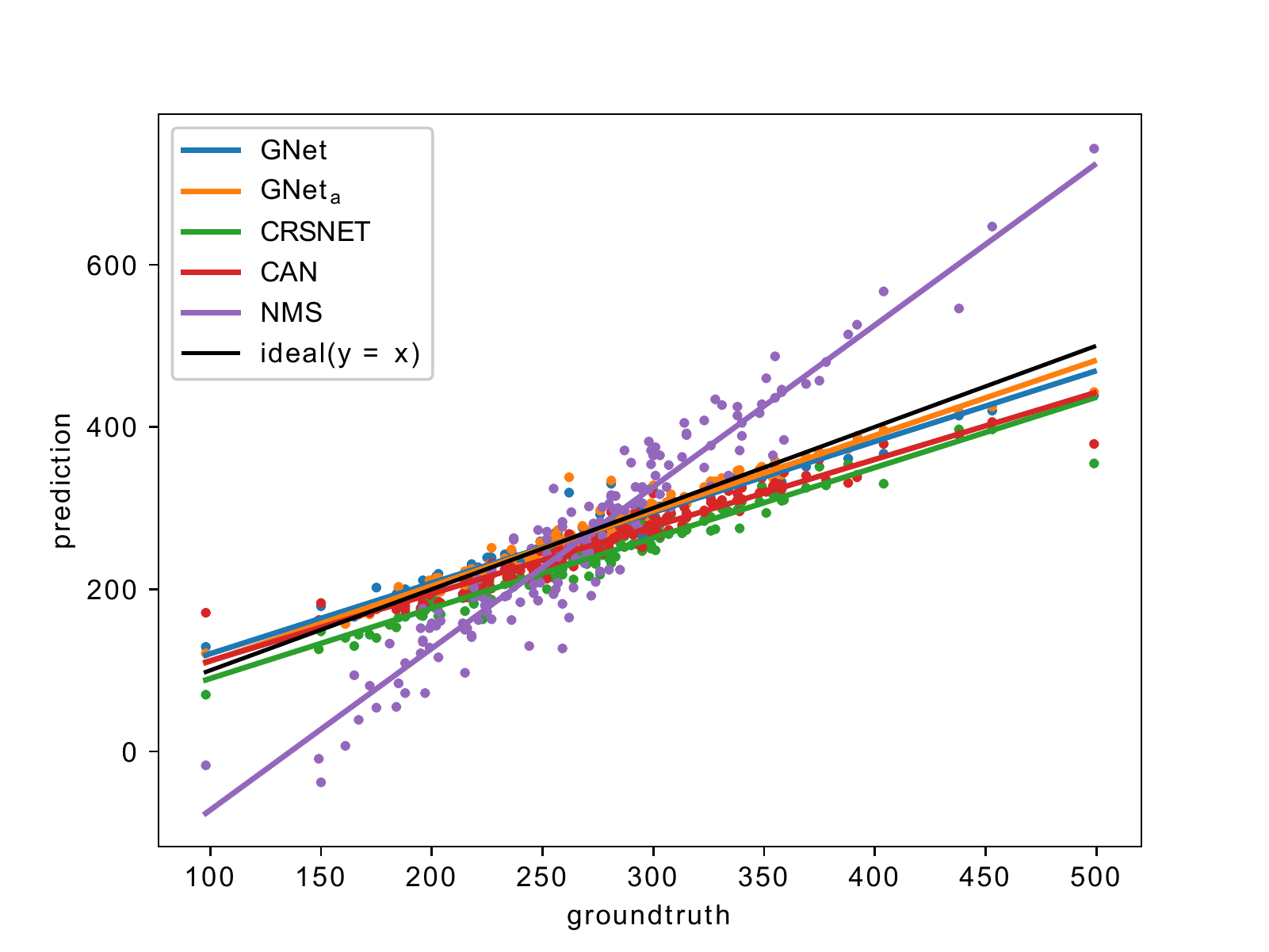} &
        \includegraphics[width=0.49\linewidth]{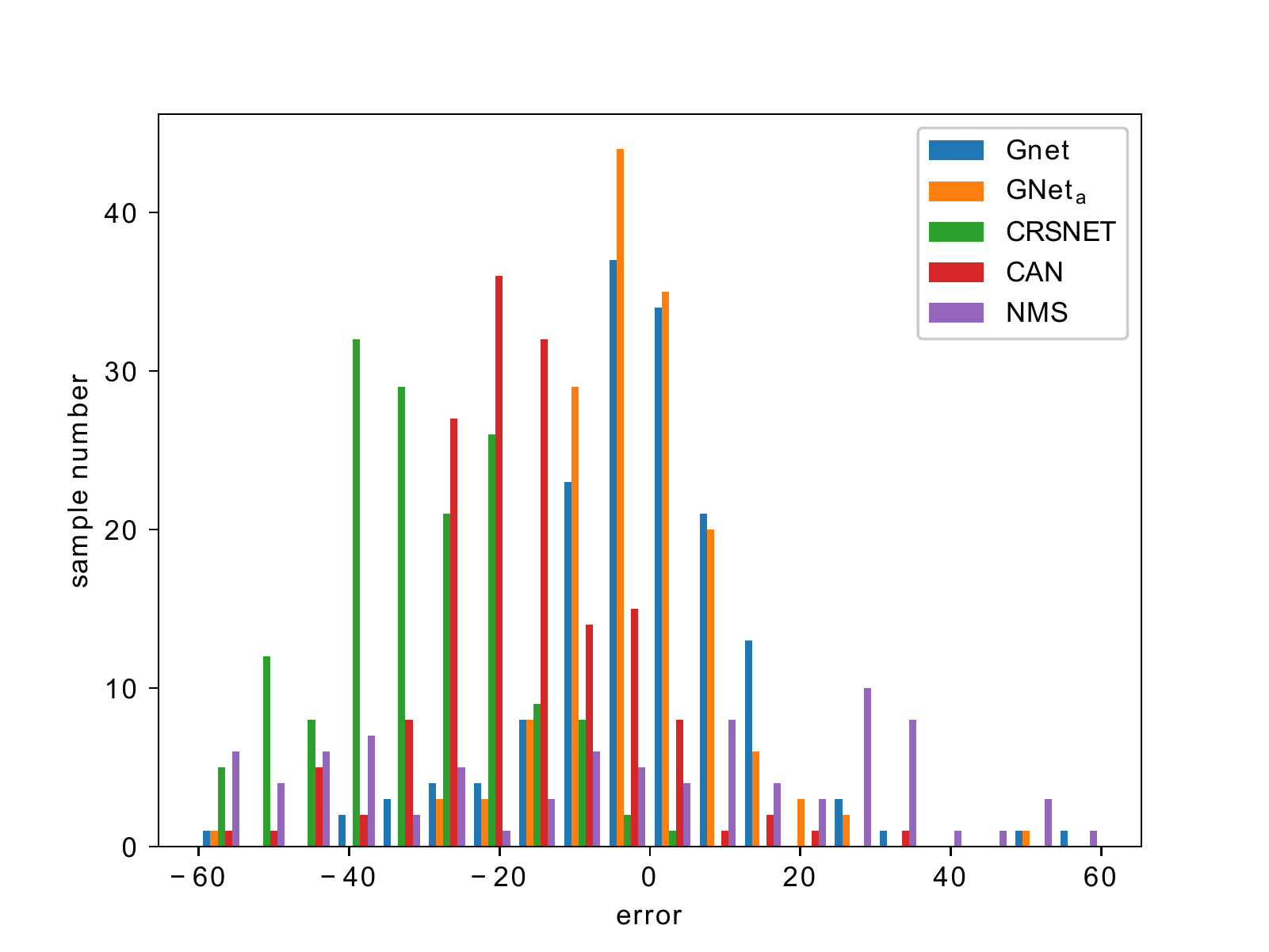} 
    \end{tabular}

\end{center}
   \caption{The linear regression between the ground truth and prediction counts for all considered methods (left) and distribution of errors (right) for $\Delta=0.5$.}
\label{fig:hist}
\end{figure}

\textbf{GNet vs. density-based methods:} The results for the four evaluated methods and two different resolutions ($\Delta=1.0$ and $\Delta=0.5$) are presented in Table~\ref{tab:datares}~left. We also add the basic non-maxima suppression method for both resolutions as a means of comparison and baseline for other methods. The GNet method outperforms both density-based methods regardless the resolution with large differences between the GNet and CSRNet/CAN especially highlighted for the lower resolution case ($\sim23$ RMSE difference). The density-based methods, when compared to GNet, tend to report fewer false positives but their relatively high count miss rate (FN) results in overall degradation in performance. This can be observed in the reported distribution of errors presented in Fig.~\ref{fig:hist} where the expected error values for CSRNet/CAN are way below the 0 mark and their distribution is more spread. Similarly, the resulting linear regression lines between the ground truth and predictions in Fig.~\ref{fig:hist} reveal a negative bias for the density-based methods although it seems that the the total count per strawberry example does not significantly influence the error rate.
High number of false negatives from NMS are explained by the non-regular geometric features of the achenes and their fusion with the flesh of the fruit. Also the numerous bumps and perturbation of the flesh (due to handling and bruising of the fruits), cause the higher number of false positive predicted by NMS compared to other methods.

The adaptive kernel variant GNet$_a$ improves results by a small margin when compared to the fixed kernel GNet, with the most significant change for the higher resolution. This indicates that adjusting the size of the kernels, especially in small and dense areas, improves the overall accuracy. There are relatively small differences between the two density-based methods, although CAN performs better in higher resolution ($\Delta=0.5$) which can be attributed to their contextual module having more pixels to work with when compared to CSRNet which is taking scales less into consideration.

\begin{table}[ht]
\begin{center}
\begin{tabular}{l | r}
    \setlength\tabcolsep{2.5pt} 
        \begin{tabular}{|c|c|c|c|c|c|c|c}
          \hline
          Arch. & $\Delta$ & RMSE   & MAE & FP\% & FN\%\\
         \hline
         GNet  & \multirow{4}{*}{0.5}      & 14.30  & 9.80  & 1.55 & 2.06\\
         GNet$_{a}$ &  & \textbf{12.86} & \textbf{8.40}  & 1.34 & \textbf{1.75}\\
         CSRNet &     & 36.42  & 32.90 & \textbf{0.00} & 12.13\\
         CAN    &      & 24.62  & 20.03 & 0.41 & 6.97\\
         NMS    &       &  71.64    &  57.37  &   11.41   &  10.45   \\
         \hline
         GNet  & \multirow{4}{*}{1.0}   & \textbf{16.10}  & \textbf{10.89} & 0.87 & 3.15\\
         GNet$_{a}$ &    & 17.60 & 10.10 & 2.65 & \textbf{1.07}\\
         CSRNet &     & 39.84  & 35.55 & \textbf{0.24}  & 12.87\\
         CAN    &     & 39.65  & 30.48 & 0.28  & 10.96\\
         NMS    &       &   71.16     &  55.75    &   7.51   &  13.74   \\
         \hline
        \end{tabular}&
        \setlength\tabcolsep{2.5pt} 
        \begin{tabular}{|c|c|c|c|c|}
              \hline
              Arch.  & MSE   & MAE & FP\% & FN\%\\
             \hline
              UNet & 13.69 & 9.21 & \textbf{0.94} & 2.45 \\
              UNet$_{t}$ &14.24 & 9.71 & 1.38 & 2.19\\
              GNet$_{a}$ & \textbf{12.86} & \textbf{8.40}  & 1.34 & \textbf{1.75}\\
              \hline
        \end{tabular}
\end{tabular}

\end{center}
\caption{Performance comparison of the GNet and density-based methods for two different resolutions $\Delta$ (left). Performance comparison for the standard UNet, UNet$_{t}$ with transpose convolutions instead of upsampling, and the proposed GNet$_{a}$ (right).}\label{tab:datares}
\end{table}

\textbf{Ablation study of GNet:} To demonstrate the benefits of the proposed improvements to the original UNet architecture, we conducted a short ablation study. We compare the performance of the standard UNet with a version replacing upsampling with transpose convolutions denoted as UNet$_{t}$, and with the proposed GNet$_{a}$ (see Table~\ref{tab:datares} right). Without dilation, replacing upsampling with transpose convolution does not bring significant improvements, while adding dilation increases the local information taken in consideration and improves the results. Slightly higher FP rates for GNet$_a$ come from difficult areas, where bumps, other defects or difficult features to annotate were highlighted, thanks to the added precision from dilation.

\begin{figure}[!ht]
\begin{center}
\begin{tabular}{l| r}
    \setlength\tabcolsep{1.5pt} 
    \begin{tabular}{c c}
    \includegraphics[width=0.1\linewidth]{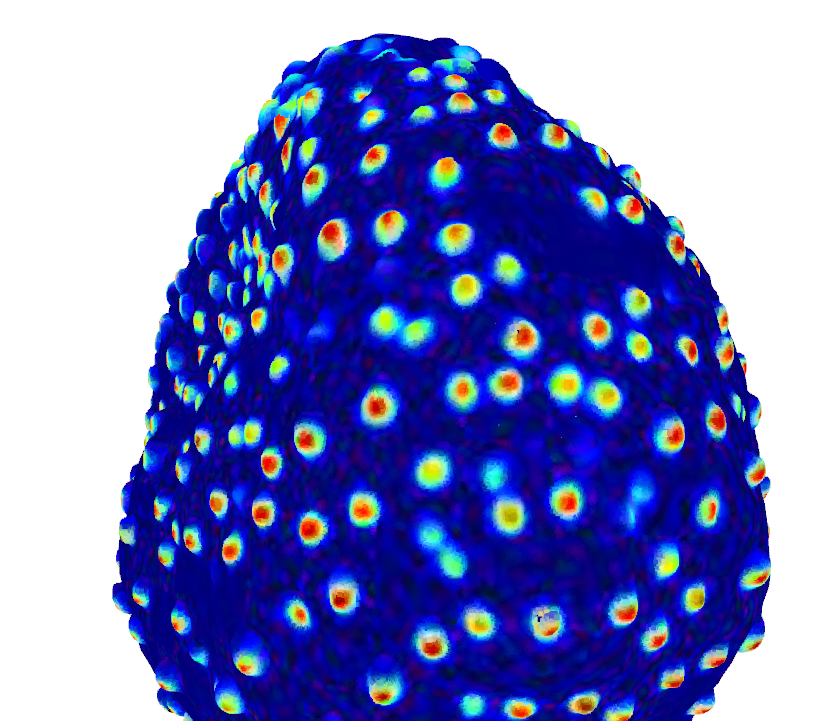}&
    \includegraphics[width=0.23\linewidth]{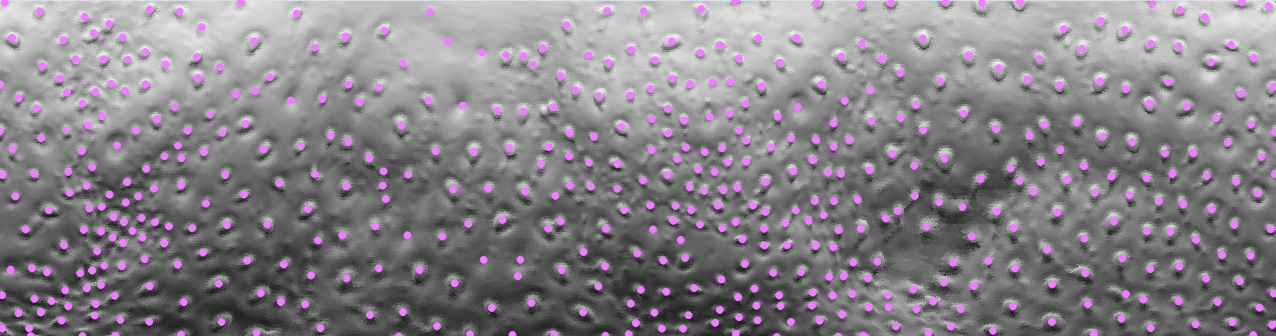}\\
    \includegraphics[width=0.23\linewidth]{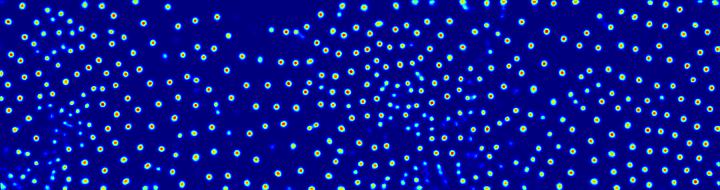}&
    \includegraphics[width=0.23\linewidth]{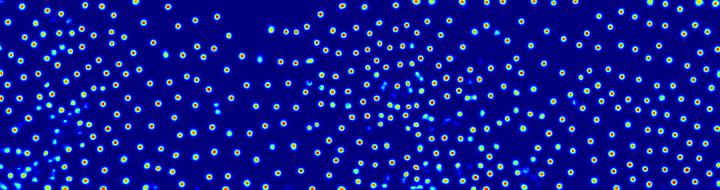}  \\
    \includegraphics[width=0.23\linewidth]{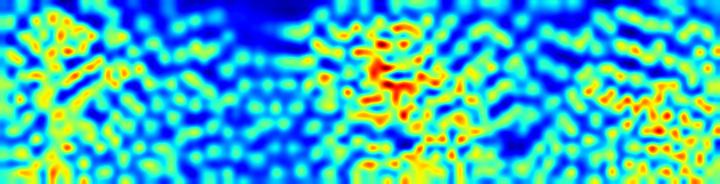}&
    \includegraphics[width=0.23\linewidth]{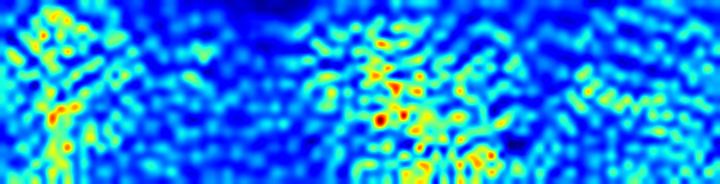} \\
    \end{tabular} &
    \setlength\tabcolsep{1.5pt} 
    \begin{tabular}{c c}
    \includegraphics[width=0.09\linewidth]{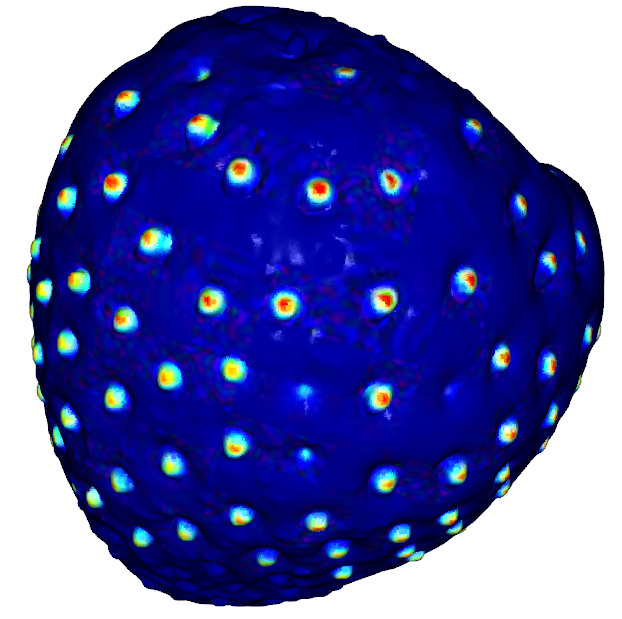}&
    \includegraphics[width=0.23\linewidth]{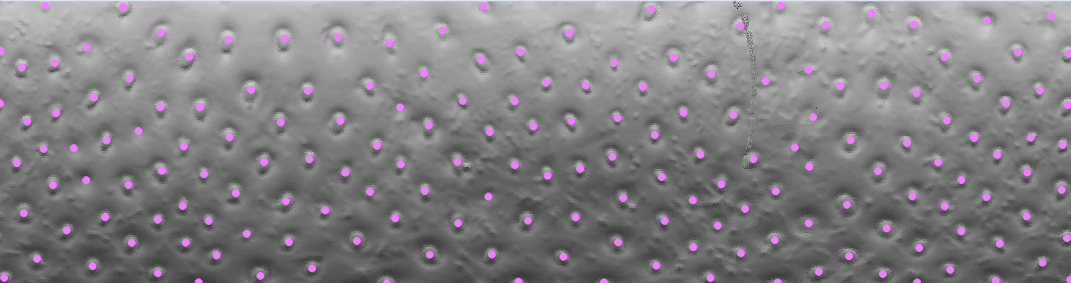}\\
    \includegraphics[width=0.23\linewidth]{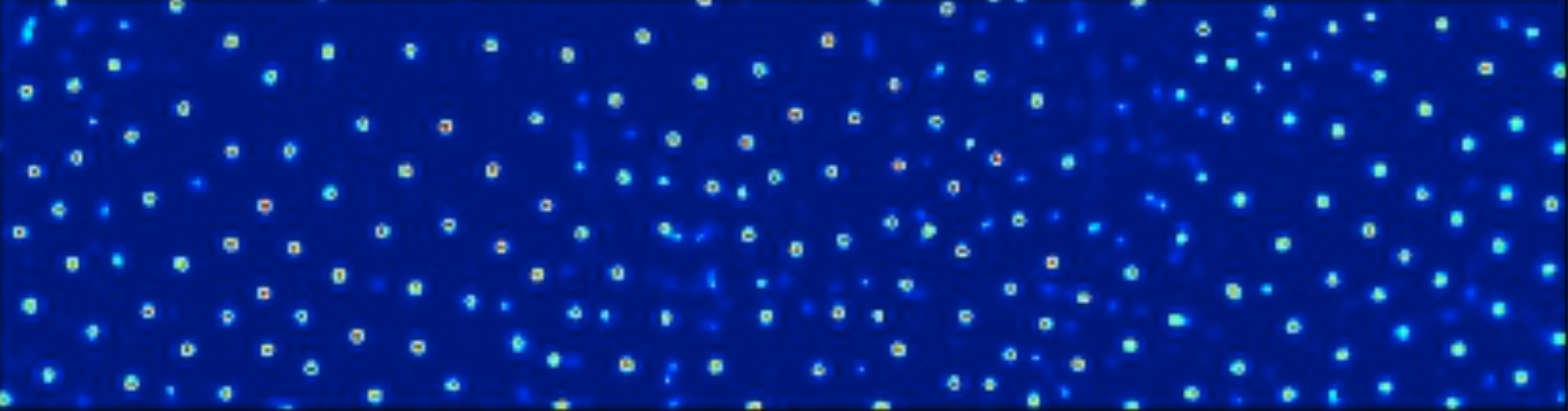}&
    \includegraphics[width=0.23\linewidth]{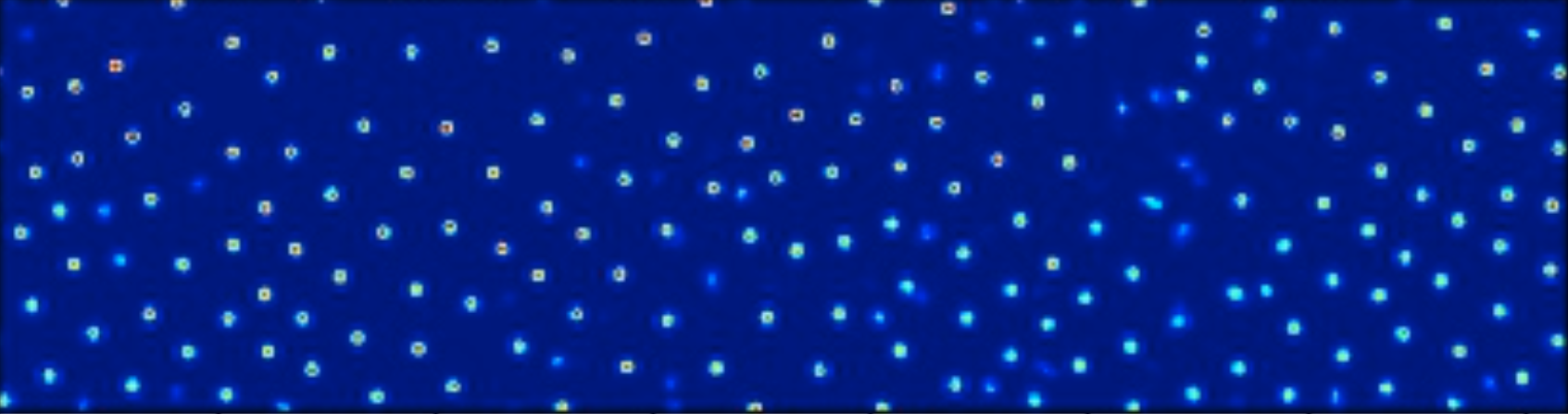}  \\
    \includegraphics[width=0.23\linewidth]{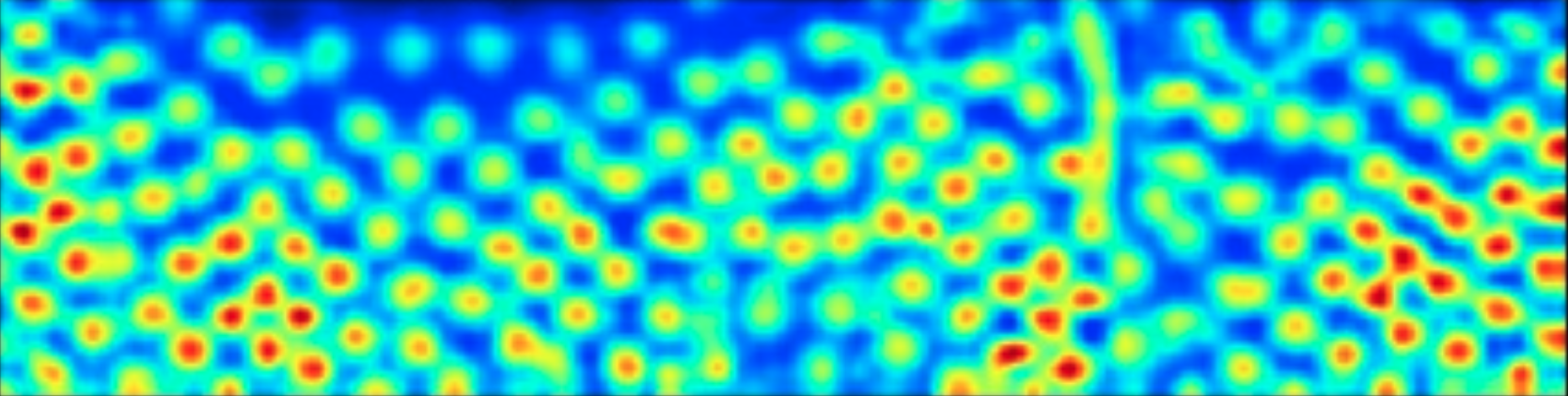}&
    \includegraphics[width=0.23\linewidth]{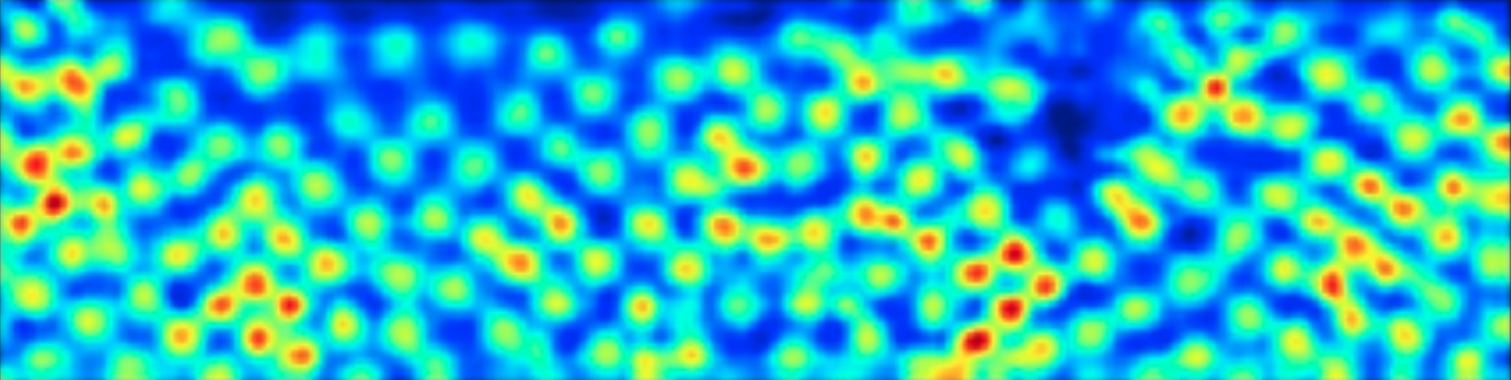} \\
    \end{tabular}
    
\end{tabular}
\end{center}
   \caption{A difficult (left) and uniform (right) example with high density areas. From left-to-right then top-to-bottom: GNet prediction over the 3D surface, annotated projection and the outputs from GNet, GNet$_{a}$, CSRNet and CAN respectively for $\Delta =0.5$.}
\label{fig:difficult}
\end{figure}

\textbf{Qualitative analysis:} We also present two graphical examples from our test set including a difficult case with very dense areas (Fig.~\ref{fig:difficult}) and an easier example with more homogeneous spread of achenes over the surface (Fig.~\ref{fig:difficult}). For both examples, it can be seen that GNet methods better predict localisation and count, with a more precise result obtained with GNet$_a$ trained with adaptive kernels.

\begin{figure}[!ht]
\begin{center}
\begin{tabular}{c c c}
    \includegraphics[width=0.22\linewidth]{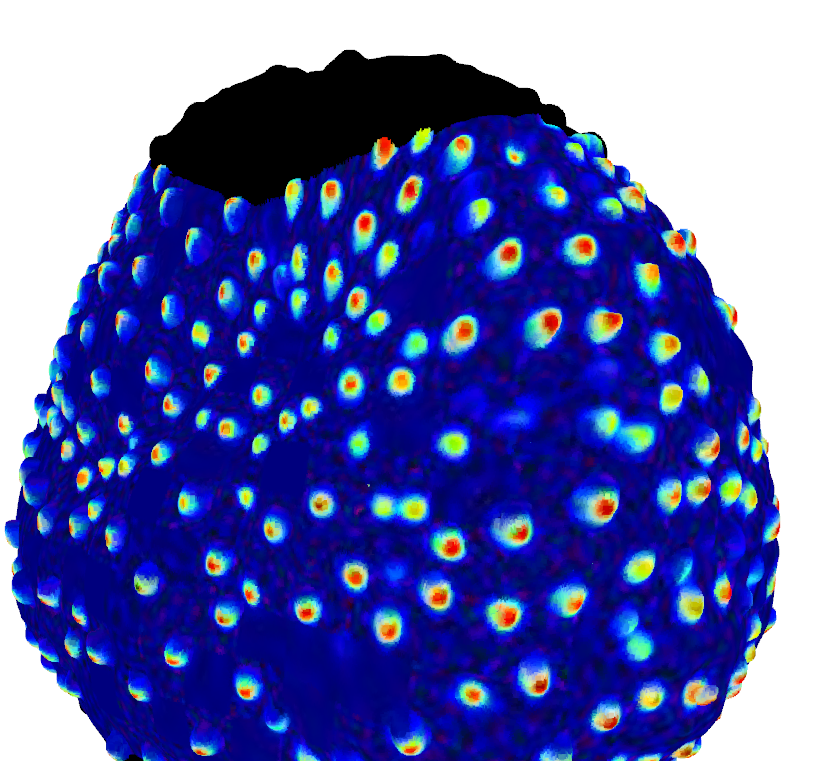}&
    \includegraphics[width=0.4\linewidth]{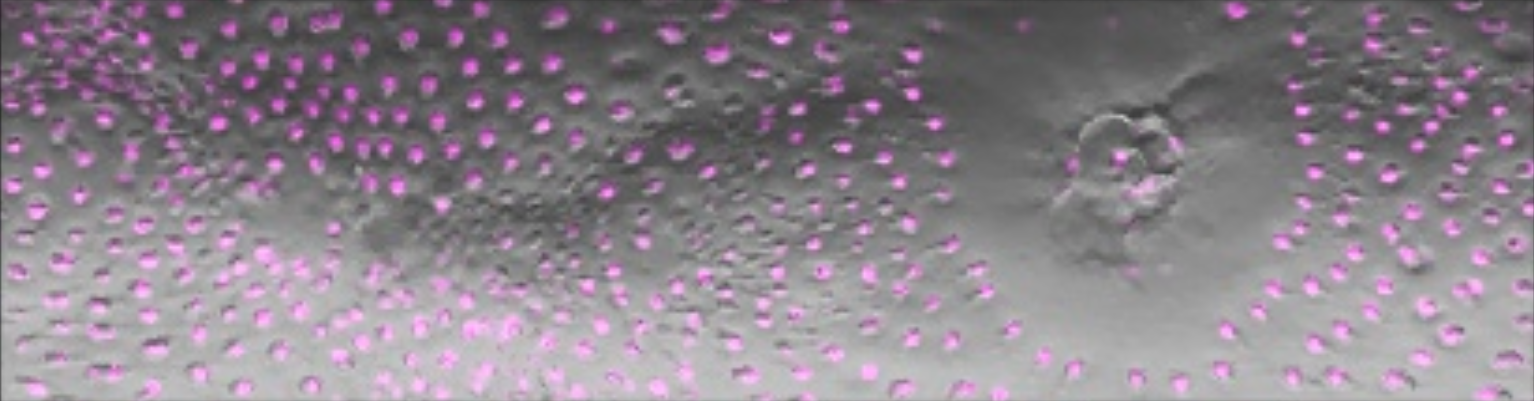}&
    \includegraphics[width=0.22\linewidth]{images/3Dsurfaceachenes.png}
    
\end{tabular}
\end{center}
  \caption{Predicting achene locations in unannotated regions: prediction in the ROI only (left) and in the unannotated bottom part of the berry on 2D projection (centre) and 3D surface (right).}
\label{fig:therest}
\end{figure}

\textbf{Unannotated regions:} We also assess if GNet$_{a}$ can generalise to unannotated achenes outside of the chosen ROI. For that purpose, we simply rotate the berry along the X axis to place its top and bottom on the central horizontal axis of the projection resulting in a 2D projection featuring previously cropped-out regions. An example result is presented in Fig.~\ref{fig:therest} where it can be seen that the trained system generalises well to the completely unseen areas of the object.

\vspace{-1em}

\section{Conclusions and Future Work}

In this paper we proposed to use spherical projection on images of spheroid objects for surface feature localisation and counting combined with Gaussian map prediction and clustering. Our method allows an accurate localisation of achenes on the surface of strawberries, outperforming state-of-the-art density-based methods for this particular application. Introducing adaptive kernels and improvements to the UNet architecture lead to even better results and extremely precise count and localisation of achenes over the 3D surface of strawberries. The presented work opens up new possibilities for exploiting surface features for other phenotyping applications considering crops with similar spheroid shapes. Future work will address issues of rotation robustness and better predictions by using different rotations of objects and their projections as well as introducing end-to-end learning for the selection of threshold locally. We are also interested in incorporating such automatic predictions into more complex phenotyping and biology pipelines, especially by correlating genotypic information with the predicted phenotypes.

\section{Acknowledgement}
This work was partially funded by the Collaborative Training Partnership for Fruit Crop Research.

\bibliography{egbib}
\end{document}